# Joint Khmer Word Segmentation and Part-of-Speech Tagging Using Deep Learning

**Rina Buoy**[†]  **Nguonly Taing**[†]  **Sokchea Kor** [‡]

[†]Techo Startup Center (TSC)

[‡]Royal University of Phnom Penh (RUPP)

`{rina.buoy,nguonly.taing}@techostartup.center`

`kor.sokchea@rupp.edu.kh`

**Abstract**

Khmer text is written from left to right with optional space. Space is not served as a word boundary but instead, it is used for readability or other functional purposes. Word segmentation is a prior step for downstream tasks such as part-of-speech (POS) tagging and thus, the robustness of POS tagging highly depends on word segmentation. The conventional Khmer POS tagging is a two-stage process that begins with word segmentation and then actual tagging of each word, afterward. In this work, a joint word segmentation and POS tagging approach using a single deep learning model is proposed so that word segmentation and POS tagging can be performed spontaneously. The proposed model was trained and tested using the publicly available Khmer POS dataset. The validation suggested that the performance of the joint model is on par with the conventional two-stage POS tagging.

## 1 Introduction

### 1.1 Background

Khmer (KHM) is the official language of the Cambodia kingdom and the Khmer script is used in the writing system of Khmer and other minority languages such Kuay, Tampuan, Jarai Krung, Brao and Kravet. Khmer language and writing system were hugely influenced by Pali and Sanskrit in early history [1].

Khmer script is believed to be originated from the Brahmi Pallava script. The Khmer writing system has been undergone ten evolutions for over 1400 years. In the early- and mid-19th century, there was an effort to romanize the Khmer language; however, it was not successful. This incident led to Khmerization in the middle of and late 19th-century [1]. Khmer is classified as a low-resource language [2].

Khmer text is written from left to right with optional space. Space is not served as a word boundary but instead, it is used for readability or other functional purposes. Therefore, word segmentation is a prior step in Khmer text processing tasks. Various efforts including dictionary-based and statistical models are made to solve the Khmer segmentation problem [3][4].

Khmer has two distinct phonological features from the other Southeast Asian languages such as Thai and Burmese. Khmer is not a tonal language and therefore, a large set of vowel phonemes are available to compensate for this absence of tones. Khmer has also a large set of consonant clusters (C1C2 or C1C2C3). Khmer allows complex initial consonant clusters at the beginning of a syllable [2].

### 1.2 POS Tagging

Part-of-Speech tagging is one of the sequence labelling tasks in which a word is assigned to one of a predefined tag set according to its syntactic function. POS tagging is required for various downstream tasks such as spelling check, parsing, grammar induction, word sense disambiguation, and information retrieval [5].

There is no explicit word delimiter in the Khmer writing system. Automatic word segmentation is run to obtain segmented words and POS tagging is performed afterwards. The performance of POS tagging is reliant on the results of segmentation in this two-stage approach [5].

For languages such as Khmer, Thai, Burmese which do not explicit word separator, the definition of words is not a natural concept and therefore, segmentation and POS tagging cannot be separated as both tasks unavoidably affect one another [2].

In this paper, we, thus, propose a joint word segmentation and POS tagging using a single deep learning network to remove the adverse dependency effect. The proposed model is a bidirectional long short-term memory (LSTM) recurrent network with one-to-one architecture. The network takes inputs as a sequence of characters and outputs a sequence of POS tags. We use the publicly available Khmer POS dataset by [5] to train and validate the model.

## 2 Related Work

### 2.1 Word Segmentation

One of the early researches on Khmer word segmentation was done by [3]. The authors proposed a bidirectional maximum matching approach (BiMM) to maximize segmentation accuracy. BiMM processed maximum matching twice – forward and backward. The average accuracy of BiMM was reported to be 98.13%. BiMM was, however, unable to handle out-of-vocabulary words(OOV) and take into account the context.

Another word segmentation approach was proposed by [6][7] which used conditional random field (CRF). The feature template was defined using a trigram model and 10 tags for each input character. 5000 sentences were manually segmented and were used to train a first CRF model which was used to segment more sentences. Additional manual hand-corrections were needed to build a training corpus of 97,340 sentences. 12,468 sentences were used as a test set. Two CRF models were trained. The 2-tags model was for predicting only word boundary and the 5-tags model for predicting both word boundary and identifying compound words. Both models obtained the same F1 score of 0.985.

[8] proposed a deep learning approach for the Khmer word segmentation task. Two long short-term memory networks were studied. The character-level network took inputs as a sequence of characters while the character-cluster networks took, instead, a sequence of character clusters (KCC).

### 2.2 Khmer POS Tagging

Khmer is a low-resource language with limited natural language processing (NLP) research. One of the earlier research on POS tagging used a rule-based transformation approach [9]. The same authors later introduced a hybrid method by combining trigram models and the rule-based transformation approach. 27 tags were defined. The dataset was manually annotated and contained about 32,000 words. For known words, the hybrid model could achieve up to 95.55% and 94.15% on training and test set, respectively. For unknown words, 90.60% and 69.84% on training and test set were obtained.

Another Khmer POS tagging study was done by [10] which used a conditional random field model. The authors recycled the tag definitions in [9] and built a training corpus of 41,058 words. The authors experimented with various feature templates by including morphemes, word-shapes and name-entities and the best template gave an accuracy of 96.38%.

Based on Choun Nat dictionary, [5] defined 24 tags, some of which were added for considering word disambiguation task. The authors used an automatic word segmentation tool by [6] to segment 12,000 sentences along with some manual corrections. Various machine learning algorithms were used to train POS tagging. These include Hidden Markov Model (HMM), Maximum Entropy (ME), Support Vector Machine (SVM), Conditional Random Fields (CRF), Ripple Down Rules-based (RDR) and Two Hours of Annotation Approach (combination of HMM and Maximum Entropy Markov Model). RDR approach outperformed the rest by achieving an accuracy of 95.33% on the test set while CRF, HMM and SVM approaches achieved comparable results.

## 2.3 Joint Segmentation and POS tagging

For most Indo-European languages, POS tagging can be done after segmentation since spaces are used to separate words in their writing system. Most Eastern and South-Eastern Asian, on the contrary, do not use any explicit word separator and the definition of words is not well defined. Segmentation and POS tagging are, therefore, cannot be separated. The authors suggested applying a joint segmentation and POS tagging for low-resource languages which share similar linguistic features as Khmer and Burmese [2].

## 3 Modified POS Tag Set

[5] proposed a comprehensive set of 24 POS tags which were derived from Choun Nat dictionary. These tag sets are shown in Table 1. In this work, we propose the following revisions to the tag set defined in [5]:

- Grouping the measure (M) tag under noun (NN) tag since the syntactic role of the measure tag is the same as the noun tag. For example, some words belonging to the measure tag are ក្បាល (head), សម្រាប់ (set), and អង្គ (person).

- Grouping the relative pronoun (RPN) tag under the pronoun tag since the relative pronoun tag has only one word (ដែល).

- Grouping the currency tag (CUR), double tag (ៗ - DBL), et cetera tag (ៗល៕ - ETC), and end-of-sentence tag (។ - KAN) under the symbol (SYM) tag.

- Grouping the injection (UH) tag under the particle tag (PA).

- Grouping the adjective verb (VB_JJ) and compliment verb (V_COM) tag under the verb (VB) tag.

After applying the above revisions, the resulting tag set consists of 15 tags which is shown in Table 2.

The descriptions of the revised POS tags are as follows:

1. Abbreviation (AB): In Khmer writing, an abbreviation can be written with or without a dot. Without an explicit dot, there is an ambiguity between a word and an abbreviation. For example, គម or គ.ម. (Kilometer).

2. Adjective (JJ): An adjective is a word used to describe a noun and is generally placed after nouns except for loanwords from Pali or Sangkrit [5]. Some common Khmer adjectives are, for example, ស (white) ល្អ (good) តូច (small), and ធំ (big). ឧត្តម and មហា are examples of Pali/Sangkrit loanword.

3. Adverb (RB): An adverb is a word used to describe a verb, adjective and another adverb [5][11]. For example, some words belonging to the adverb tag are ពេក (very), ណាស់ (very), ហើយ (already), and ទើប (just).

4. Auxiliary Verb (AUX): Only three words are tagged as an auxiliary verb and their syntactic role is to indicate tense [5]. បាន or មាន indicates past tense. កំពុង indicates progress tense. នឹង indicates future tense.

5. Cardinal Number (CD): A cardinal number is used to indicate the quantity [5]. Some examples of cardinal numbers are ១ (one), បី (three), ៤ន (four), and លាន (million).

6. Conjunction (CC): A conjunction is a word used to connect words, phrases or clauses [5][11]. For example, some words belonging to the conjunction tag are បើ (if), ប្រសិនបើ (if), ពីព្រោះ (because), ព្រោះ (because), and ប៉ុន្តែនោះសោត (nevertheless).

7. Determiner Pronoun (DT): A determiner is a word used to indicate the location or uncertainty of a noun. Determiners are equivalent to English words: this, that, those, these, all, every, each, and some. In Khmer grammar, a determiner is tagged as either a pronoun or adjective [11]. However, a determiner pronoun tag is used in [5] and this work. For example, some words belonging to the determiner pronoun tag are នេះ (this), នោះ (that), សព្វ (every), ទាំងនេះ (these),    (those), and ខ្លះ (some).

Table 1. The POS Tag Set Proposed by [5]

| No. | Tags | | No. | Tags | |
|---|---|---|---|---|---|
| 1 | AB | Abbreviation | 13 | NN | Noun |
| 2 | AUX | Auxiliary Verb | 14 | PN | Proper Noun |
| 3 | CC | Conjunction | 15 | PA | Particle |
| 4 | CD | Cardinal Number | 16 | PRO | Pronoun |
| 5 | CUR | Currency | 17 | QT | Question Word |
| 6 | DBL | Double Sign | 18 | RB | Adverb |
| 7 | DT | Determiner Pronoun | 19 | RPN | Rel. Pronoun |
| 8 | ETC | Khmer Sign | 20 | SYM | Symbol/Sign |
| 9 | IN | Preposition | 21 | UH | Interjection |
| 10 | JJ | Adjective | 22 | VB | Verb |
| 11 | KAN | Khmer Sign | 23 | VB_JJ | Adjective Verb |
| 12 | M | Measure | 24 | V_COM | Verb Compliment |

Table 2. The Revised POS Tag Set Proposed in This Work

| No. | Tags | | No. | Tags | |
|---|---|---|---|---|---|
| 1 | AB | Abbreviation | 9 | NN | Noun |
| 2 | AUX | Auxiliary Verb | 10 | PN | Proper Noun |
| 3 | CC | Conjunction | 11 | PA | Particle |
| 4 | CD | Cardinal Number | 12 | PRO | Pronoun |
| 5 | DT | Determiner Pronoun | 13 | QT | Question Word |
| 6 | IN | Preposition | 14 | RB | Adverb |
| 7 | JJ | Adjective | 15 | SYM | Symbol/Sign |
| 8 | VB | Verb | | | |

8. Pronoun (PRO): A pronoun is a word used to refer to a noun or noun phrase which was already mentioned [5][11]. In this work, a pronoun tag is used to tag both a personal pronoun and a relative pronoun. In Khmer grammar, there is only one relative pronoun which is ដែល (that, which, where, who).

9. Preposition (IN): A preposition is a word used with a noun, noun phrase, verb or verb phrase to time, place, location, possession and so on [5][11]. For example, some words belonging to the preposition tag are នៅលើ (above), កាលពី (from), តាម (by), and អំពី (about).

10. Noun (NN): A noun is a word used to identify a person, animal, tree, place, and object in general [5][11]. For example, some words belonging to the noun tag are សិស្ស (student), គោ (cow), តុ (table), and កសិករ (farmer).

11. Proper Noun (PN): A proper noun is a word used to identify the name of a person, animal, tree, place, and location in particular [5][11]. For example, some words belonging to the proper noun are សុខា (Sokha - a person's name), ភ្នំពេញ (Phnom Penh), កម្ពុជា (Cambodia), and ស៊ីធីអិន (CTN).

12. Question Word (QT): Some examples of question word are តើ (what) and ដូចម្ដេច (how).

13. Verb (VB): A verb is a word used to describe action, state, or condition [5][11]. For example, some words belonging to the verb tag are ដើរ (to walk), ជា (to be), មើល (to watch), and ស្រេក (to be thirsty). [5] invented two more tags for verb, which were the adjective verb (VB_JJ) and compli-ment verb (V_COM). A VB_JJ was used to tag any verb behaving like an adjective in large compound words such as ម៉ាស៊ីនបោកខោអាវ (washing machine), កំបិតចិតបន្លែ (knife) and so on. A V_COM, on the other hand, was used to tag any verb in verb phrases or collocations such as រៀនចេះ (to learn), ប្រលងជាប់ (to pass), and so on. Both VB_JJ and V_COM are dropped in this work for the reason that identifying the VB_JJ and V_COM should be done via compound and collocation identification task [12]. Certain Khmer compounds are loose and have the same structure as a Subject-Verb-Object sentence [13][2]. This is illustrated in Figure 1.

14. Particle (PA): There is no clear concept of a particle in Khmer grammar. [5] identified three groups of particles namely: hesitation, response and final. Some examples of particles are អឺ (hesitation), អើ (response), សិន (final), and ចុះ (final)។

15. Symbol (SYM): Various symbols in writing are grouped under the SYM tag. The SYM tag includes currency signs, Khmer signs (។, ៕, ៗ) and various borrowed signs ( +, -, ?) and so on.

## 4  POS Tagging Methodology

The application of deep learning in the Khmer word segmentation task was first proposed by [8]. In this work, we extend [8] by combining word segmentation and POS tagging task in a single deep learning model. The proposed model is a variant of recurrent neural networks. The details are explained below.

### 4.1  Recurrent Neural Network

A recurrent neural network (RNN) is a type of neural network with a cycle in its connections. That means the value of an RNN cell depends on both inputs and its previous outputs. Elman networks or simple recurrent networks have been proven to be very effective in NLP tasks [14]. An illustration of a simple recurrent network is given in Figure 2.

The hidden vector, $h_t$ depends on both input, $x_t$ and previous hidden vector, $h_{t-1}$.

$$h_t = g(Uh_{t-1} + Wx_t) \quad (1)$$

Where:

- $U$ and $W$ are trainable weight matrices. They are shared by all time steps in the sequence.

The fact that the computation of $h_t$ requires $h_{t-1}$ makes RNN an ideal candidate for sequence tagging tasks such as POS tagging.

### 4.2  Recurrent Neural Network for Sequence Labelling

Various forms of RNN architecture are given in Figure 3. The choice of an RNN architecture depends on the application of interest. A POS tagging task is a sequence labelling and one-to-one architecture is suitable. A possible RNN model for POS tagging is given in Figure 4. Here are the forward steps taken by an RNN model:

1. A sequence of inputs are encoded or embedded with an embedding layer.

2. Hidden vectors are computed by unrolling the computational graph through time.

3. At each time step, the RNN cell outputs an output vector.

4. A Softmax activation is applied to the output vectors.

### 4.3  Stacking

One or more RNN layers can be stacked on top of each other to form a deep RNN model. An illustration of a stacked RNN model is shown in Figure 4. In a stacked RNN model, an input sequence is fed to the first RNN layer to produce $h_t^1$. $h_t^1$ is then fed to another RNN layer until the layer output layer.

Stacking is used to learn representations at different levels across layers. The optimal number of the stack depends on the application of interest and is a matter of hyperparameter tuning [14].

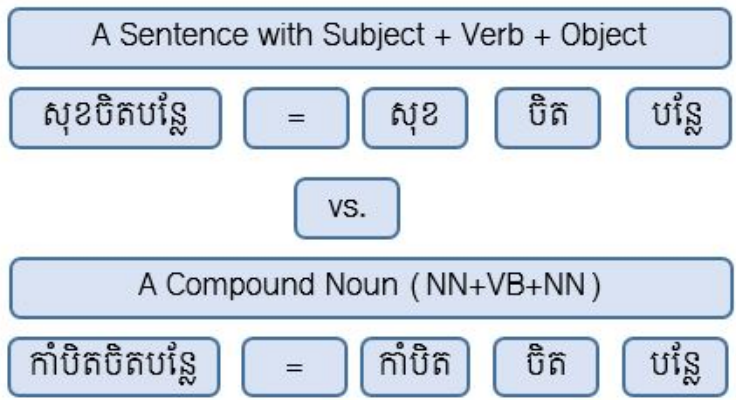

Figure 1. Similar Structure of a Sentence and Compound Noun

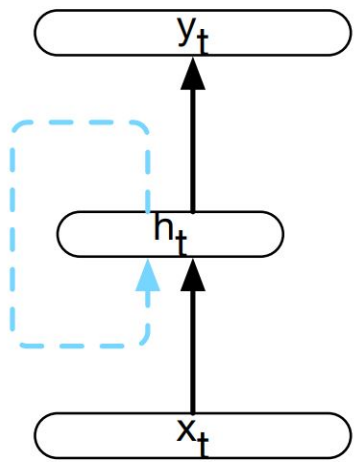

Figure 2. A simple RNN [14]

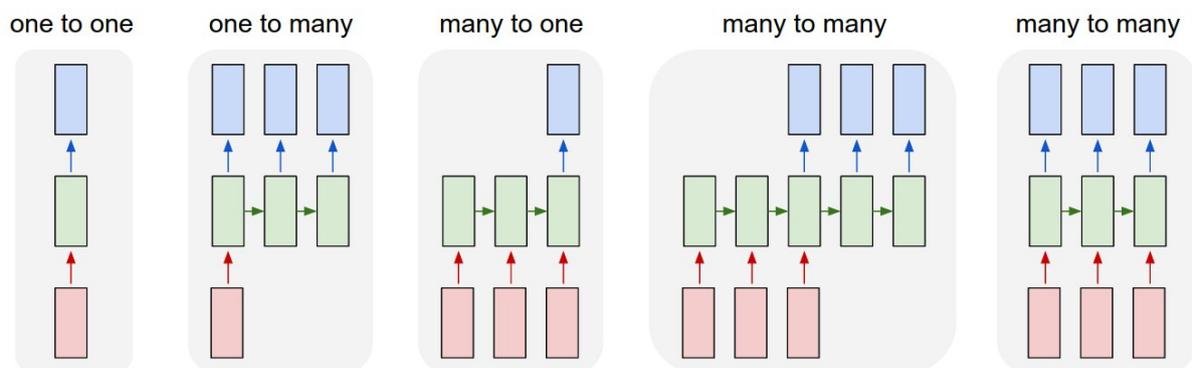

Figure 3. Various RNN architectures [15]

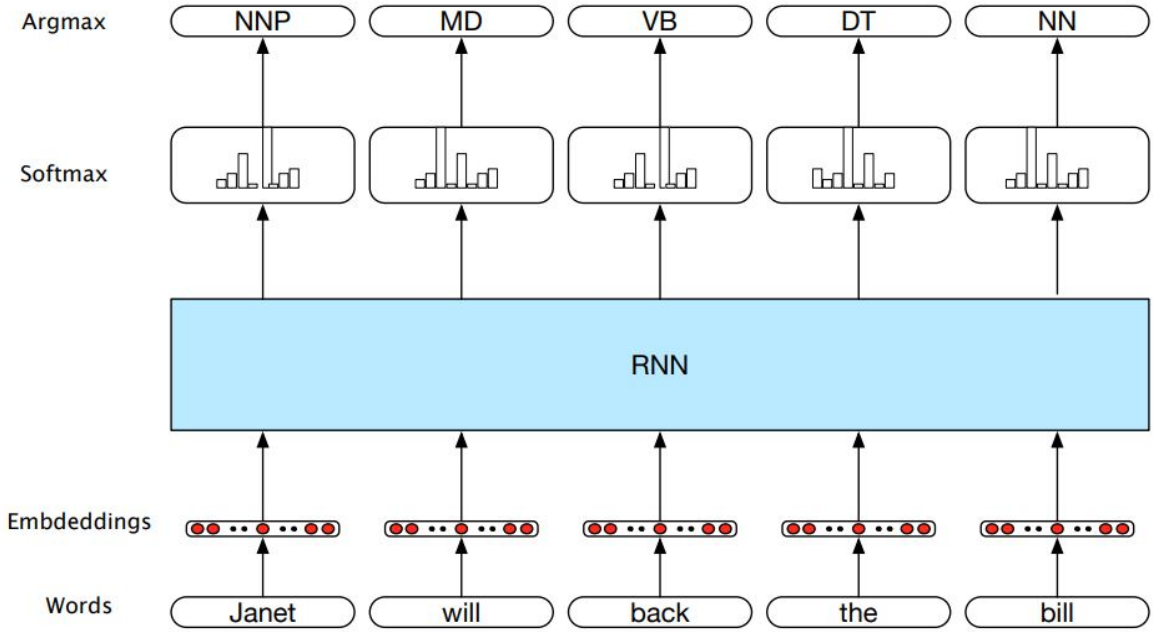

Figure 4. A example RNN model for POS tagging task [14]

### 4.4 Bidirectionality

In forward mode (from left to right), $h_t^1$ represents what the model has processed from the first input in the sequence until the input at a time, $t$.

$$h_t^f = RNN_{forward}(x_1^t) \qquad (2)$$

Where:

- $h_t^f$ represents the forward hidden vector after the network sees an input sequence up to time step, t.

- $x_1^t$ are inputs up to time, $t$.

In certain applications such as sequence labelling, a model can have access to the entire input sequence. It is possible to process an input sequence from both directions - forward and backward [14]. $h_t^b$ representing the hidden vector up to time, $t$ in reverse order can be expressed as:

$$h_t^b = RNN_{backward}(x_t^n) \qquad (3)$$

Where:

- $h_t^b$ represents the backward hidden vector after the network sees an input sequence from timestep, n to t.

- $x_t^n$ are inputs up to time, $t$ in reserve order.

An RNN model which processes from both directions is known as a bidirectional RNN (bi-RNN). $h_t^f$ and $h_t^b$ can be averaged or concatenated to form $h_t$.

### 4.5 Long Short-Term Memory (LSTM)

When processing a long sequence, an RNN cell has difficulty in carrying forward critical information for two reasons [14]:

- The weight matrices need to provide information for the current output as well as the future outputs.

- The back-propagation through time suffers from vanishing gradients problem due to repeated multiplications along a long sequence.

Long Short-Term Memory (LSTM) networks were devised to address the above issues by introducing sophisticated gate mechanisms and an additional context vector to control the flow of information into and out of the units [14]. Each gate in LSTM networks consists of:

1. A feed-forward layer
2. A sigmoid activation
3. Element-wise multiplication with the gated layer

The choice of a sigmoid function is to gate information flow as it tends to squash the output to either zero or one. The combined effect of a sigmoid activation and element-wise multiplication is the same as binary masking [14].

There are three gates in an LSTM cell. The details are as follow:

- Forget Gate: As the name implies, the objective of a forget gate is to remove irrelevant information from the context vector. The equations of a forget gate are given below:

$$f_t = \sigma(U_f h_{t-1} + W_f x_t) \quad (4)$$
$$k_t = c_{t-1} \odot f_t \quad (5)$$

  Where:
  - $\sigma$ is a sigmoid activation.
  - $U_f$ and $W_f$ are weight matrices.
  - $f_t$ and $c_t$ are vectors at time, $t$.
  - $c_{t-1}$ is a previous context vector.
  - $\odot$ is an element-wise multiplication operator.

- Add Gate: As the name implies, the objective of an add gate is to add relevant information to the context vector. The equations of an add gate are given below:

$$g_t = tanh(U_g h_{t-1} + W_g x_t) \quad (6)$$

  Like a standard RNN cell, the above equation extracts information from the previous hidden vector and current input.

$$i_t = \sigma(U_i h_{t-1} + W_i x_t) \quad (7)$$
$$j_t = g_t \odot i_t \quad (8)$$

  The current context vector is updated as follows:

$$c_t = j_t + k_t \quad (9)$$

  Where:
  - $U_i$, $W_i$, $U_g$, and $W_g$ are weight matrices.
  - $g_t$, $i_t$, and $j_t$ are vectors at time, $t$.
  - $c_t$ is a current context vector at time, $t$.

- Output Gate: As the name implies, the objective of an output gate is to determine what information is needed to update a current hidden vector at time, $t$:

$$o_t = \sigma(U_o h_{t-1} + W_o x_t) \quad (10)$$
$$h_t = o_t \odot tanh(c_t) \quad (11)$$

  Where:
  - $U_o$ and $W_o$ are weight matrices.
  - $o_t$ is a vectors at time, $t$.
  - $h_t$ is a hidden vector at time, $t$.

### 4.6 Bidirectional LSTM Network for Joint Word Segmentation and POS Tagging

In this section, we introduce a bidirectional LSTM (Bi-LSTM) network at character level for joint word segmentation and POS tagging. It is bidirectional since the model has access to an entire input sequence during forward run.

The descriptions of the proposed Bi-LSTM network shown in Figure 5 are as follows:

1. Inputs: The network takes a sequence of characters as inputs.
2. Inputs Encoding: One-hot encoding is used to encode an input character.
3. LSTM Layers: The network processes an input sequence in both directions - forward and backward. The forward and backward hidden vector in the final LSTM stack is concatenated to form a single hidden vector.
4. Feed-forward Layer: The concatenated hidden vector is fed into a feed-forward layer to produce an output vector. The size of the output vector is equal to the number of POS tags plus one as an additional no-space (NS) tag is introduced. No space tag is explained in the output representation section.

5. Softmax: Softmax activation is applied to the output vector to produce probabilistic outputs.

In the proposed models of [8], a sigmoid function is used to output a probability whether a character or cluster is the starting of a word.

### 4.7 Cross Entropy Loss

Multi-class cross-entropy loss is used to train the proposed model. The loss is expressed as follows:

$$L(y, \hat{y}) = \sum_{k=1}^{K} y^k log(\hat{y}^k) \qquad (12)$$

Where:

- $\hat{y}^k$ is a predicted probability of class, $k$.
- $y^k$ is 0 or 1, indicating whether class, $k$ is the correct classification.

## 5 Experimental Setup

### 5.1 Dataset - Train/Test Set

The dataset used in this work is obtained from [5]. The dataset is available on Github (https://github.com/ye-kyaw-thu/khPOS) under CC BY-NC-SA 4.0 license. The dataset consists of 12,000 sentences (25,626 words in total). The average number of words per sentence is 10.75. The word segmentation tool from [6] was used and manual corrections were also performed. The original dataset has 24 POS tags. The most common tags are NN, PN, PRO, IN and VB.

We revised the dataset by grouping some tags as per the above discussion. The revised dataset has 15 tags in total - 9 tags fewer. The count and proportion of each tag are given in Figure 3 in descending order.

The 12,000-sentence dataset was used as a training set. [5] provided a separate open test set for evaluating the model performance.

### 5.2 Inputs and Labels Preparation

An example of a training sentence is given below.

ខ្ញុំ/PRO ស្រលាញ់/VB ខ្មែរ/PN ។/SYM
(I love Khmer)

Table 3. POS Tag Frequency in the Training Set

| Tags | Frequency | Percentage |
|------|-----------|------------|
| NN | 32297 | 25.03% |
| PN | 20084 | 15.57% |
| VB | 18604 | 14.42% |
| PRO | 13950 | 10.81% |
| IN | 13446 | 10.42% |
| RB | 6428 | 4.98% |
| SYM | 5839 | 4.53% |
| JJ | 4446 | 3.45% |
| DT | 4311 | 3.34% |
| CD | 3337 | 2.59% |
| CC | 2788 | 2.16% |
| AUX | 2466 | 1.91% |
| PA | 885 | 0.69% |
| QT | 79 | 0.06% |
| AB | 69 | 0.05% |
| Total | 129029 | 100.00% |

- Space denotes word boundary.
- POS tag of a word is right after slash.

The corresponding input and target sequence of the above training sentence is illustrated in Figure 6. If a character is the starting character of a word, its label is the POS tag of the word and it also means the beginning of a new word. Otherwise, the no-space (NS) tag is assigned.

### 5.3 Training Configuration

The network was implemented in the PyTorch framework and trained on Google Colab Pro. Training utilized mini-batch on GPU.

The following hyper-parameters were used:

- Number of LSTM stacks = 2
- Hidden dimension = 100
- Batch size = 128
- Optimizer = Adam with learning rate of 0.001
- Epochs = 100
- Loss function = categorical cross entropy loss

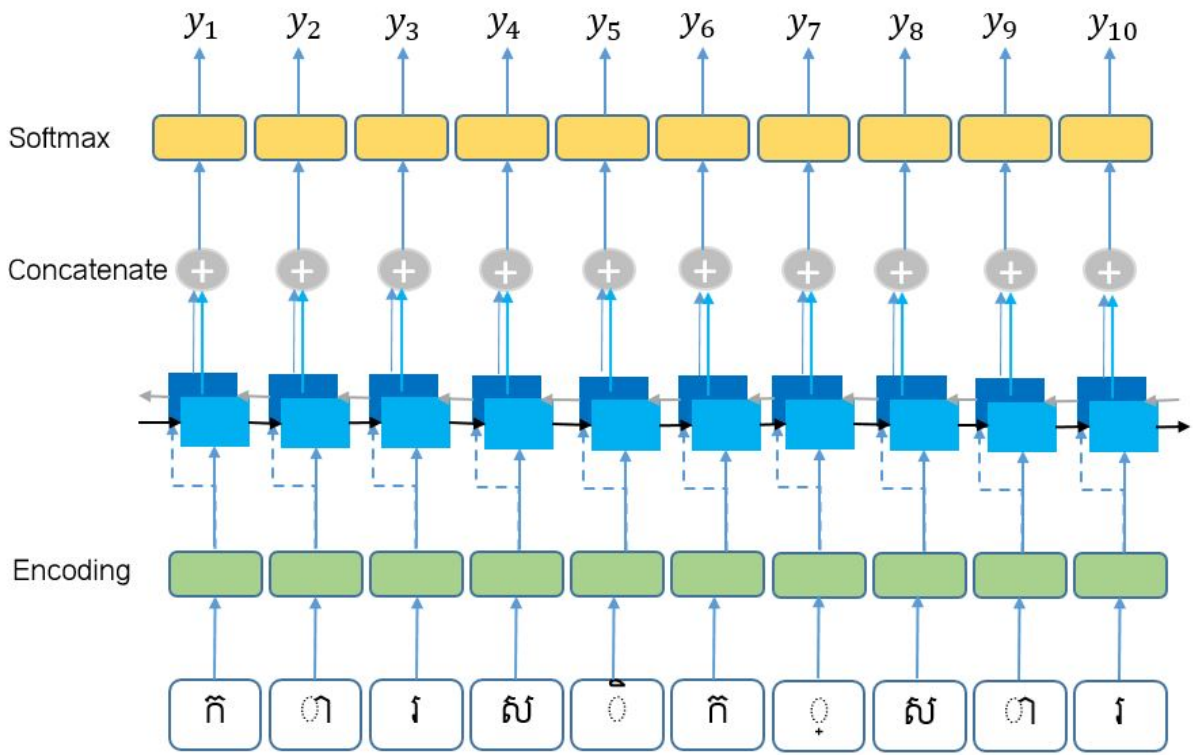

Figure 5. The proposed Bi-LSTM network for joint segmentation and POS tagging at character level

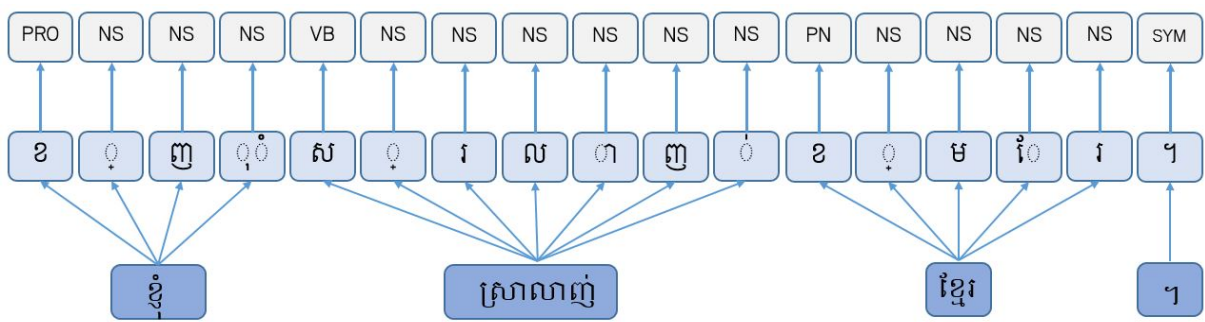

Figure 6. Input and Output Sequence Representation

Table 4. Accuracy of Word Segmentation

| Metric | Training Set | Test Set |
|---|---|---|
| Accuracy | 99.27% | 97.11% |

Each input character was encoded to a one-hot vector of 132 dimensions, which is the number of Khmer characters including numbers and other signs.

## 6 Results and Evaluation

The accuracy of word segmentation is defined as:

$$Accuracy = \frac{Count_{word}^{correct}}{Count_{word}^{corpus}} \quad (13)$$

Where:

- $Count_{word}^{correct}$ is the number of correctly segmented words.
- $Count_{word}^{corpus}$ is the number of works in the corpus.

The accuracy for a given tag, $i$ is defined as below:

$$Accuracy = \frac{Count_{pos,i}^{correct}}{Count_{pos,i}^{corpus}} \quad (14)$$

Where:

- $Count_{pos,i}^{correct}$ is the number of correctly predicted POS tag, $i$.
- $Count_{pos,i}^{corpus}$ is the number of POS tag, $i$ in the corpus.

The overall POS tagging accuracy is given by:

$$Accuracy = \frac{\sum_{i=1}^{tag} Count_{pos,i}^{correct}}{\sum_{i=1}^{tag} Count_{pos,i}^{corpus}} \quad (15)$$

Where:

- $tag$ is the number of POS tags.

The segmentation and overall POS tagging accuracy are given Table 4 and 6, respectively while the accuracy breakdowns by tag are given in Table 5

Table 5. POS Tag Accuracy Breakdowns

| Tags | Train | Test |
|---|---|---|
| NN | 98.43% | 93.75% |
| PN | 98.94% | 96.40% |
| VB | 96.90% | 89.12% |
| PRO | 98.86% | 97.39% |
| IN | 97.49% | 92.05% |
| RB | 95.00% | 87.23% |
| SYM | 100.00% | 99.74% |
| JJ | 92.35% | 79.89% |
| DT | 99.71% | 96.99% |
| CD | 98.52% | 96.82% |
| CC | 96.06% | 94.09% |
| AUX | 100.00% | 98.15% |
| PA | 95.77% | 83.33% |
| QT | 100.00% | 100.00% |
| AB | 100.00% | 83.33% |

Table 6. Overall Accuracy of POS tagging

| Metric | Training Set | Test Set |
|---|---|---|
| Accuracy | 98.14% | 94.00% |

## 7 Discussion

The trained model achieved the segmentation and POS tagging accuracy of 97.11% and 94.00%, respectively. Compared with [5], our POS tagging accuracy was 1.33% lower on the open test set. In the work of [5], the overall error should be composed of two components - segmentation and POS tagging error and is approximated by the below equation:

$$\epsilon_t = \epsilon_s + \epsilon_p \quad (16)$$

Where:

- $\epsilon_t$: is the overall error.
- $\epsilon_s$: is the segmentation error.
- $\epsilon_p$: is the POS tagging error.

Since the segmentation error was not included in [5], the reported error was just the POS tagging error. [5] used the segmentation tool by [6] with the reported error ($\epsilon_s$) of 1.5%. The estimated overall error ($\epsilon_t$) of [5] is about 6.17% since the highest accuracy of [5] was reported to be 95.33% ($\epsilon_p = 4.67\%$).

Thus, the performance of the joint segmentation and POS tagging model with $\epsilon_t$ of

6.00% is on par with the conventional two-stage POS tagging method.

# 8 Conclusion and Future Work

In this work, we proposed joint word segmentation and POS tagging using a deep learning approach. We presented a bidirectional LSTM network that takes inputs at the character level and outputs a sequence of POS tags. The overall accuracy of the proposed model is on par with the conventional two-stage Khmer POS tagging. We believe the training dataset available is limited in size and a significantly larger dataset is required to train a more robust joint POS tagging model with greater generalization ability.

# References

[1] Makara Sok. *Phonological Principles and Automatic Phonemic and Phonetic Transcription of Khmer Words.* PhD thesis, Payap University, 2016.

[2] Chenchen Ding, Masao Utiyama, and Eiichiro Sumita. Nova: A feasible and flexible annotation system for joint tokenization and part-of-speech tagging. *ACM Trans. Asian Low-Resour. Lang. Inf. Process.*, 18(2), December 2018.

[3] Narin Bi and Nguonly Taing. Khmer word segmentation based on bi-directional maximal matching for plaintextand microsoft word. *Signal and Information Processing Association Annual Summit and Conference (APSIPA)*, 2014.

[4] Chea Sok Huor, Top Rithy, Ros Pich Hemy, Vann Navy, Chin Chanthirith, and Chhoeun Tola. Word bigram vs orthographic syllable bigram in khmer word. *PAN Localization Team*, 2007.

[5] Ye Kyaw Thu, Vichet Chea, and Yoshinori Sagisaka. Comparison of six pos tagging methods on 12k sentences khmer language pos tagged corpus. *1st Regional Conference on OCR and NLP for ASEAN Languages*, 2017.

[6] Vichet Chea, Ye Kyaw Thu, Chenchen Ding, Masao Utiyama, Andrew Finch, and Eiichiro Sumita. Khmer word segmentation using conditional random fields. *Khmer Natural Language Processing*, 2015.

[7] Ye Kyaw Thu, Vichet Chea, Andrew Finch, Masao Utiyama, and Eiichiro Sumita. A large-scale study of statistical machine translation methods for khmer language. In *Proceedings of the 29th Pacific Asia Conference on Language, Information and Computation*, pages 259–269, Shanghai, China, October 2015.

[8] Rina Buoy, Nguonly Taing, and Sokchea Kor. Khmer word segmentation using bilstm networks. *4th Regional Conference on OCR and NLP for ASEAN Languages*, 2020.

[9] C. Nou and W. Kameyama. Khmer pos tagger: A transformation-based approach with hybrid unknown word handling. In *International Conference on Semantic Computing (ICSC 2007)*, pages 482–492, 2007.

[10] Sokunsatya Sangvat and Charnyote Pluempitiwiriyawej. Khmer pos tagging using conditional random fields. *Communications in Computer and Information Science*, 2017.

[11] National Council of Khmer Language. *Khmer Grammar Book.* National Council of Khmer Language, 2018.

[12] Wirote Aroonmanakun. Thoughts on word and sentence segmentation in thai. 2007.

[13] Sok Khin. *Khmer Grammar.* Royal Academy of Cambodia, 2007.

[14] Dan Jurafsky and James H. Martin. *Speech and Language Processing.* 3rd ed draft, 2020.

[15] Andrej Karpathy. The unreasonable effectiveness of recurrent neural networks.